\documentclass[letterpaper, 10 pt, conference]{ieeeconf}

\IEEEoverridecommandlockouts
\usepackage{graphicx}
\usepackage{url}
\usepackage{stfloats}
\usepackage[tight]{subfigure}
\usepackage{amsmath,bm}
\usepackage{gensymb}
\usepackage{amssymb}
\usepackage{amsfonts}   
\usepackage{amsopn}     
\usepackage{psfrag}
\usepackage{mdwtab}     
\usepackage{enumerate}
\usepackage{mdwlist}
\usepackage{accents}
\usepackage{enumerate}
\usepackage{mdwtab}     
\usepackage{mdwlist}
\usepackage{color}
\usepackage{algorithmicx}
\usepackage{algorithm} 
\usepackage{algpseudocode} 
\usepackage{acronym}
\usepackage{paralist}
\usepackage[english]{babel}
\usepackage[utf8]{inputenc}

\acrodef{COM}[\textsc{COM}]{Center of Mass}
\acrodef{COP}[\textsc{COP}]{Center of Pressure}
\acrodef{DOF}[\textsc{DOF}]{Degrees of Freedom}
\acrodef{RRT}[\textsc{RRT}]{Rapidly-exploring Random Tree}
\acrodef{LIP}[\textsc{LIP}]{Linear Inverted Pendulum}
\acrodef{CBF}[\textsc{CBF}]{Control Barrier Function}
\acrodef{HZD}[\textsc{HZD}]{Hybrid Zero Dynamics}
\acrodef{MPC}[\textsc{MPC}]{Model Predictive Control}
\acrodef{ZMP}[\textsc{ZMP}]{Zero Moment Point}
\acrodef{DoF}[\textsc{DoF}]{Degrees of Freedom}
\acrodef{DCBF}[\textsc{DCBF}]{Discrete \ac{CBF}}
\acrodef{OSC}[\textsc{OSC}]{Operational Space Control}
\acrodef{QP}[\textsc{QP}]{Quadratic Program}
\acrodef{CLF}[\textsc{CLF}]{Control Lyapunov Function}
\acrodef{WBC}[\textsc{WBC}]{Whole-Body Controller}
\acrodef{CWC}[\textsc{CWC}]{Contact Wrench Cone}

\newtheorem{rem}{\bf Remark}

\newcommand{\Qi}[1]{ q^\iota_{\mathrm{#1}} }

\title{\LARGE \bf
A Sequential MPC Approach to Reactive Planning for Bipedal Robots
}
\author{Kunal S. Narkhede, Abhijeet M. Kulkarni, Dhruv A. Thanki and Ioannis Poulakakis
\thanks{This work is supported in part by NSF MRI-2018905.}
\thanks{K.S Narkhede, A. M. Kulkarni, D. A. Thanki and I. Poulakakis are with Department of Mechanical Engineering,
        University of Delaware, Newark, Delaware 19716, USA.
        {\tt\small \{kunalnk, amkulk, thankid and poulakas\}@udel.edu}}%
}

\begin{document}
\addtolength{\topmargin}{.15in} 

\maketitle
\pagestyle{empty}

\begin{abstract}
This paper presents a sequential Model Predictive Control (MPC) approach to reactive motion planning for bipedal robots in dynamic environments. The approach relies on a sequential polytopic decomposition of the free space, which provides an ordered collection of mutually intersecting obstacle-free polytopes and waypoints. These are subsequently used to define a corresponding sequence of MPC programs that drive the system to a goal location avoiding static and moving obstacles. This way, the planner focuses on the free space in the vicinity of the robot, thus  alleviating the need to consider all the obstacles simultaneously and reducing computational time. We verify the efficacy of our approach in high-fidelity simulations with the bipedal robot Digit, demonstrating robust reactive planning in the presence of static and moving obstacles.
\end{abstract}

\section{Introduction}
\label{sec:intro}

\begin{figure}[t]
\vskip +2pt
\centering
{
\includegraphics[width=0.99\columnwidth]{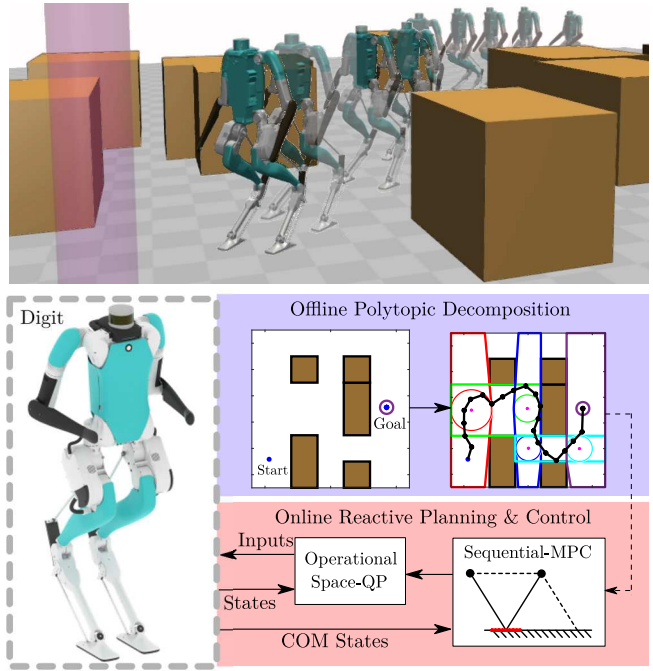}
}
\vskip -15pt
\caption{Top: The bipedal robot Digit navigating amidst obstacles. Bottom: Overview of the proposed framework. An RRT-guided sequential decomposition of the workspace to obstacle-free polytopes is used to formulate a sequence of MPC programs that drive the LIP to the goal position and avoid collisions with static and moving obstacles. The suggested LIP-based plan is then realized on Digit using a QP-based operational space controller.}
\label{fig:overview}
\vskip -15pt
\end{figure}

Navigating amidst obstacles is a prerequisite for bringing robots into human-populated spaces. Owing to their morphological and functional traits, humanoids and bipedal robots like Digit (cf. Fig.~\ref{fig:overview}) are ideally suited for such spaces. To bring such robots a step closer to navigating in cluttered environments, this paper proposes a sequential \ac{MPC} approach, which enables reactive trajectory planning and realization so that static and moving obstacles in the robot's vicinity can be avoided. 

Several methods for planning and executing walking motions on bipedal robots exist in the literature~\cite{Harada_2010_book}; many rely on \emph{reduced-order models}~\cite{Wieber2016ModelingAC, Wensing2017BLL} to resolve the complexity of the full dynamics of such robots. Reducing the robot's dynamics to a lower-dimensional subsystem can be done---for example---by imposing virtual constraints using feedback~\cite{AmesPoulakakis2017BLL}; these subsystems can be used to define a library of Dynamic Movement Primitives for sampling-based motion planning~\cite{Mohamad2016CDC, Veer2017IROS}. Although these methods offer formal stability guarantees~\cite{Veer2019ICRA,Veer2019TAC, Veer2020TAC}, the resulting reduced-order models are in general hard to interpret intuitively; however, see~\cite{Gong2021Arxiv} for recent results on this matter. An alternative approach---and the one we adopt in this paper---is to select an intuitive  reduced-order model \emph{a priori}, and then use it to plan desired trajectories for the robot. Prominent among such models for bipedal walking is the \ac{LIP}~\cite{Kajita2001IROS}, which has been instrumental in relating \ac{COM} desired trajectories to sequences of footstep locations in an online~\cite{Ijspeert2014RSS, Tsagarakis2019IROS} and robust~\cite{Kheddar2019ICRA} fashion. 

These references are but a few instances of a broader set of gait generation methods that use the \ac{LIP} as a predictive model in the context of \ac{MPC} schemes~\cite{Wieber2006Humanoids}. Many more variants of such schemes can be found in the literature, with recent results addressing the fundamental issue of stability and recursive feasibility~\cite{Oriolo2020TRO}.
The flexibility afforded by the \ac{MPC} formulation allows for the inclusion of distance-based constraints to ensure both trajectory tracking and obstacle avoidance~\cite{Soueres2017RAL, Xiong2021ICRA}. 
However, constraints based on distance are activated only when the robot comes close to an obstacle~\cite{Zeng2021ACC}. An alternative approach relies on the notion of a discrete-time \ac{CBF}, which has been introduced in~\cite{Agrawal2017RSS} to avoid collisions between a bipedal walker and obstacles in its environment.
Recently, discrete-time \ac{CBF}s have been used within \ac{MPC} to expand the nodes of a \ac{RRT} to plan paths that ensure static obstacle avoidance for the \ac{LIP}~\cite{Ghaffari2021Arxiv}; the suggested plans were then realized in the bipedal robot Cassie. 
In both~\cite{Agrawal2017RSS, Ghaffari2021Arxiv}, one \ac{CBF} constraint is introduced per static obstacle, which may result in unnecessarily large optimization programs for highly cluttered spaces.

A \ac{LIP}-based plan of \ac{COM} and footstep locations specifies the desired motion of different parts of the robot in the Cartesian space. Thus, it defines distinct control objectives, which can be realized via \ac{OSC} with different priority levels~\cite{Sentis2010TRO, Wieber2016ModelingAC}. Optimization-based techniques can be used to formulate the problem as a \ac{QP} with various control objectives expressed as equality and inequality constraints~\cite{fujimoto1998IROS}. The resulting programs can be solved using strict or weighted prioritization. In the former, lower priority objectives are achieved unless higher priority ones are compromised; this can be done by solving a hierarchy of \ac{QP}s~\cite{wensing2013ICRA, Fourquet2013TRO}, or a single \ac{QP} with additional pre-computations~\cite{Escande2014IJRR, Kim2018IROS}. Weighted prioritization, on the other hand, considers the most critical control objectives as constraints with the remaining ones weighted into a single performance index~\cite{Ijspeert2014RSS, Kheddar2019ICRA}.

In this paper, we propose a weighted \ac{QP} formulation akin to~\cite{mordatch2010ACM} to realize a \ac{LIP}-based collision-free plan computed in an \ac{MPC} fashion. Rather than incorporating static obstacles by using one \ac{CBF} per obstacle---as is commonly done in the relevant literature, e.g.~\cite{Agrawal2017RSS, Ghaffari2021Arxiv}---we propose an \emph{RRT-guided sequential decomposition} of the robot's environment in obstacle-free polytopes that connect the starting and goal locations. These polytopes capture the free space \emph{locally}, in the vicinity of the robot, 

thus alleviating the need to consider all the obstacles---including those far from the robot---simultaneously, at each solution step. The end result is a corresponding sequence of \ac{MPC} programs that drives the robot to the goal while keeping it within the free polytopes. 
Because of the significant benefits in terms of computational time, moving obstacles can be incorporated in our formulation and the associated \ac{MPC}s can be solved multiple times within each step of Digit to enhance robustness. 

The method is implemented in high-fidelity simulations with Digit, demonstrating robust realization of \ac{LIP}-based motion plans in the presence of static and moving obstacles.

\section{A 3D LIP Model for Bipedal Motion Planning}
\label{sec: 3DLIP}

\begin{figure}[b]
\vskip -10pt
\centering
{
\includegraphics[width=0.97\columnwidth]{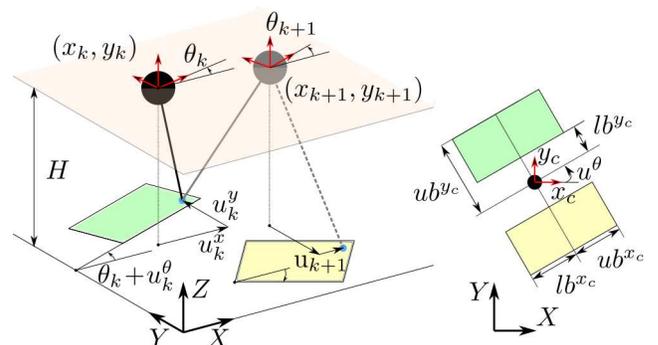}
}
\vskip -10pt
\caption{Reachable spaces for the left (green) and right (yellow) feet, defined with bounds relative to the frame $\{x_c, y_c, z_c\}$ attached at the COM. Left and right rectangles are symmetric about the $x_c$ axis when $u^\theta=0$.}
\label{fig:3Dlip}
\end{figure}

In its common configuration, the 3D \ac{LIP} comprises a point mass atop a massless telescopic leg (cf. Fig.~\ref{fig:3Dlip}). Let $(x,y) \in \mathbb{R}^2$ be the location of the mass with respect to an inertia frame and $(\dot{x}, \dot{y}) \in \mathbb{R}^2$ the corresponding velocity. It is assumed that the mass is constrained to move in a horizontal plane located at constant height $H$. We use $u^x$ and $u^y$ to denote the distance of the stance foot from the COM in the $x$ and $y$ directions, respectively (cf. Fig.~\ref{fig:3Dlip}). 

The continuous-time evolution of the 3D \ac{LIP} in the forward direction---that is, along the $x$ axis---is governed by $\ddot{x} = - (g / H) u^x$, where $g$ is the gravitational acceleration. This expression can be integrated in closed form to result in 
\begin{equation} \label{eq:LIP_x_continuous}
 \begin{bmatrix} x(t) \\ \dot{x}(t) \end{bmatrix} = 
 \begin{bmatrix} 1 & \frac{1}{\omega}\sinh{\omega t} \\ 0 & \cosh{\omega t} \end{bmatrix}
 \begin{bmatrix} x(0) \\ \dot{x}(0) \end{bmatrix} +
 \begin{bmatrix} 1-\cosh{\omega t} \\ -\omega \sinh{\omega t} \end{bmatrix}
 u^x
\end{equation}
where $x(0)$, $\dot{x}(0)$ are initial conditions and $\omega = \sqrt{g/H}$. The motion along the $y$ axis is governed by identical dynamics.

We are interested in using the 3D \ac{LIP} to plan motions for Digit. However, the robot's ability to realize the suggested motions is limited by its physical constraints. These include the joint limits of the legs and the free space on the ground available for foot placement. To incorporate these constraints, we approximate the reachable region of the swing foot with a rectangle as in~\cite{Ghaffari2021Arxiv} (cf. Fig.~\ref{fig:3Dlip}). The orientation of this region relative to the vertical axis of a body-fixed frame $\{x_c, y_c, z_c\}$ is denoted by $\theta$. We assume that $\theta$ is updated instantaneously at the exchange of support and that $\dot{\theta}=0$ during a step. Thus, if $\theta(0)$ is the angle of the rectangle of the leg providing support, the corresponding angle for the swing foot when it becomes the support foot in the forthcoming step is 
\begin{equation}\label{eq:LIP_theta_continuous}
\theta(t) = \theta(0) + u^\theta
\end{equation}
where $u^\theta$ is a step change assumed to satisfy $lb^\theta \leq u^\theta \leq ub^\theta$ for suitable lower and upper bounds $lb^\theta$ and $ub^\theta$.

Next, we assume that each step has duration $T$ and that $[k T, (k+1) T)$ is the interval spanned by the $k$-th step. 
If 
\begin{equation} \nonumber
    \mathrm{x}_k = 
    \begin{bmatrix}
    x_k & \dot{x}_k & y_k & \dot{y}_k & \theta_k
    \end{bmatrix}^\mathsf{T} 
    ~~\text{and}~~
    \mathrm{u}_k = 
    \begin{bmatrix}
    u^x_k & u^y_k & u^\theta_k
    \end{bmatrix}^\mathsf{T}    
\end{equation}
denote the state and input at the beginning of the $k$-th step, the state of the system at the $(k+1)$ step is given by
\begin{equation}\label{eq:LIP}
    \mathrm{x}_{k+1} = A \mathrm{x}_k + B \mathrm{u}_k 
\end{equation}
where $A = \mathrm{diag}\{ \hat{A},\ \hat{A}, \ 1 \}$ and $B = \mathrm{diag}\{\hat{B}, \ \hat{B}, \ 1 \}$ are block diagonal matrices with
\begin{equation} \nonumber
    \hat{A} =  
    \begin{bmatrix}
    1 & \frac{1}{\omega} \sinh{\omega T} \\ 0 & \cosh{\omega T}
    \end{bmatrix}
    ~\text{and}~
    \hat{B} = 
    \begin{bmatrix}
    1 - \cosh{\omega T} \\ -\omega \sinh{\omega T}
    \end{bmatrix} .
\end{equation}
We will be interested in planning \ac{LIP} motions in the $(x,y)$ plane. Thus, we associate with \eqref{eq:LIP} the output equation 
\begin{equation}\label{eq:LIP_output}
    \mathrm{y}_k = C \ \mathrm{x}_k
\end{equation}
where $C$ is the $2 \times 5$ matrix that extracts from $\mathrm{x}$ the position $(x,y)$ of the \ac{COM} of the 3D-\ac{LIP}; that is, $C \mathrm{x} = \begin{bmatrix} x & y \end{bmatrix}^\mathsf{T}$.  

To avoid 3D-\ac{LIP} motions that challenge the capabilities of the robot, we incorporate constraints in the \ac{LIP} step-by-step update dynamics \eqref{eq:LIP}-\eqref{eq:LIP_output}. This can be done via the following time-dependent constraint set for the state and inputs
\begin{align}
    \nonumber
    \mathcal{XU}_k = \big\{ (\mathrm{x}_k, \mathrm{u}_k)  ~|~ \mathrm{lb}_k &\leq R(\theta_k + u^\theta_k)^\mathsf{T} \mathrm{u}_k \leq \mathrm{ub}_k, \\
    \label{eq:constraint_set}
     ~\text{and}~ \delta_\mathrm{min} &\leq \sqrt{\Delta x_k^2 + \Delta y_k^2} \leq \delta_\mathrm{max} \big\}
\end{align}
the first part of which corresponds to constraints on the reachability rectangles and their orientation, while the second part to constraints on the COM's travel distance. In \eqref{eq:constraint_set}, $\mathrm{lb}_k = \begin{bmatrix} lb^{x_c}_k & lb^{y_c}_k & lb^{\theta}_k\end{bmatrix}^\mathsf{T}$ (resp. $\mathrm{ub}_k$) includes lower (resp. upper) bounds of the reachability constraints and $R(\theta_k + u^\theta_k)$ is the corresponding 3D rotation matrix (cf. Fig.~\ref{fig:3Dlip}).
In addition, $\delta_\mathrm{min}$ and $\delta_\mathrm{max}$ are the lower and upper bounds on the travel distance of the COM between steps, and $\Delta x_k = x_{k+1}-x_k$ and $\Delta y_k = y_{k+1}-y_k$. This constraint is added to avoid infeasible motions and bound the average speed.
Appendix~\ref{app:LIP-constraints} provides more details on the \ac{LIP} constraints.

\section{LIP Reactive Planning via MPC}
\label{sec:DCBF_MPC}

This section presents a method to reactively plan a path that takes the \ac{LIP} from an initial position to a goal position and avoids obstacles in the workspace $\mathcal{W}$.

\subsection{RRT-Guided Decomposition of the Free Space}

Consider first the static part $\mathcal{W}^\mathrm{s}$ of the workspace. We ``inflate'' the obstacles to account for the nontrivial dimensions of Digit, which is assumed to be enclosed in a disc of radius $0.5 \text{m}$. We also assume that all the obstacles forming $\mathcal{W}^\mathrm{s}$ are convex. While this assumption does not entail significant loss of generality\footnote{Non-convex obstacles can be treated as suggested in~\cite{Deits2015_IRIS} by decomposing them into convex pieces.}, it will allow us to compute efficiently a sequence of free polytopes $\mathcal{H}_i$, $i = 0,1,...,M-1$, along a pre-computed obstacle-free path, and use them to steer the 3D-\ac{LIP} to the goal via a chain of \ac{MPC} programs. 

To do this, we begin with generating a sequence of points in $\mathcal{W} \setminus \mathcal{W}^\mathrm{s}$ that connect an initial location with a final desired one. This can be done using any sampling-based planning algorithm; here, we use $\text{RRT}^*$ to find a sequence of such points $\Pi = \{\mathrm{p}_1,...,\mathrm{p}_K\}$ in $\mathcal{W}\setminus \mathcal{W}^{\rm s}$, which leads to the goal. While generating $\Pi$, we make sure that the line segment joining any two consecutive points in $\Pi$ does not intersect any static obstacle. With the availability of $\Pi$, we implement Algorithm~\ref{alg:RRT} to construct a chain of free polytopes and obtain a collection of waypoints to be used in a sequence of \ac{MPC} programs that connect the initial location to the goal. Algorithm~\ref{alg:RRT} calls the following functions.

\begin{algorithm}[b!]
\caption{PolyFsGen algorithm}\label{alg:RRT}
\begin{algorithmic}[1]
\State Given: $\mathrm{y}_0$, $\Pi=\{\mathrm{p}_1,..., \mathrm{p}_K\}$, $\mathrm{w}_{\rm g}$, $\mathcal{W}^{\rm s}$
\textbf{\State $\mathcal{H}_0$ = $\mathsf{Generate\_Polytopes}( \mathcal{W}^{\rm s},\mathrm{y}_0)$}
\State Initialize list: $i = 0$, $j = 0$, $\mathcal{G} = \{(\mathcal{H}_0, \mathrm{y}_0)\}$
\State $\mathrm{p}_0 = \mathrm{y}_0$, $\mathrm{p}_{K+1} = \mathrm{w}_{\rm g}$ 
        \While{$\mathrm{w}_{\rm g} \notin \mathcal{H}_i$ OR $j < K+1$}
        \State $j\gets j+1$
        \If{$\mathrm{p}_j \in \mathcal{H}_{i}$}
            \State \textbf{continue}
        \Else
            \State $\mathcal{H}_{\rm new} \gets \mathsf{Generate\_Polytopes}(\mathcal{W}^{\rm s}, \mathrm{p}_j)$
            \State $\mathcal{\bar{G}} = 
                \mathsf{Intersecting\_Polytopes}(\mathcal{H}_i,\mathrm{p}_{j-1},\mathcal{H}_{\rm new},\mathrm{p}_{j})$
                \State $\mathcal{G} \gets \mathcal{G}\cup\mathcal{\bar{G}}$
                \State $i \gets \mathsf{length}(\mathcal{G})$
                \State $\mathcal{H}_i \gets \mathcal{H}_{\rm new}$
        \EndIf
    \EndWhile
\State \Return $\mathcal{G}$
\State
\Function{$\mathsf{Intersecting\_Polytopes}$}{$\mathcal{H}_{i_1},\mathrm{p}_{j_1},\mathcal{H}_{i_2},\mathrm{p}_{j_2}$}
\State ${\rm w}_{\rm new} = \mathsf{Chebyshev\_Center}(\mathcal{H}_{i_1},\mathcal{H}_{i_2})$
\If{${\rm w}_{\rm new}$ is empty}
    \State $\ell = 0$, $\mathrm{p}^{0} = \mathrm{p}_{j_1}$, $\mathcal{H}^{0} = \mathcal{H}_{i_1}$
    \While{$\mathrm{w}_{\rm new}$ is empty}
        \State $\mathrm{p}^{\ell+1} = \mathsf{Poly\_Line\_Intersect}(\mathcal{H}^{\ell}, \mathrm{p}^{\ell},\mathrm{p}_{j_2})$
        \State $\mathcal{H}^{\ell+1} = \mathsf{Generate\_Polytopes}(\mathcal{W}^{\rm s}, \mathrm{p}^{\rm \ell+1})$
        \State ${\rm w}^{\ell+1} = \mathsf{Chebyshev\_Center}(\mathcal{H}^{\ell},\mathcal{H}^{\ell+1})$
        \State ${\rm w}_{\rm new} =  \mathsf{Chebyshev\_Center}(\mathcal{H}^{\ell+1},\mathcal{H}_{i_2})$
        \State $\ell \gets \ell+1$
    \EndWhile
\State $\mathcal{\bar{G}} \gets \{(\mathcal{H}^{1},{\rm w}^1),...,(\mathcal{H}^{\ell},{\rm w}^{\ell}),(\mathcal{H}_{i_2},{\rm w}_{\rm new}) \}$
\Else
    \State $\mathcal{\bar{G}} \gets \{(\mathcal{H}_{i_2},{\rm w}_{\rm new})\}$
\EndIf
\State \Return $\mathcal{\bar{G}}$
\EndFunction
\end{algorithmic}
\end{algorithm}

\begin{figure}[t!]
\vskip +2pt
\centering
\subfigure[]
{
\includegraphics[width=0.22\textwidth]{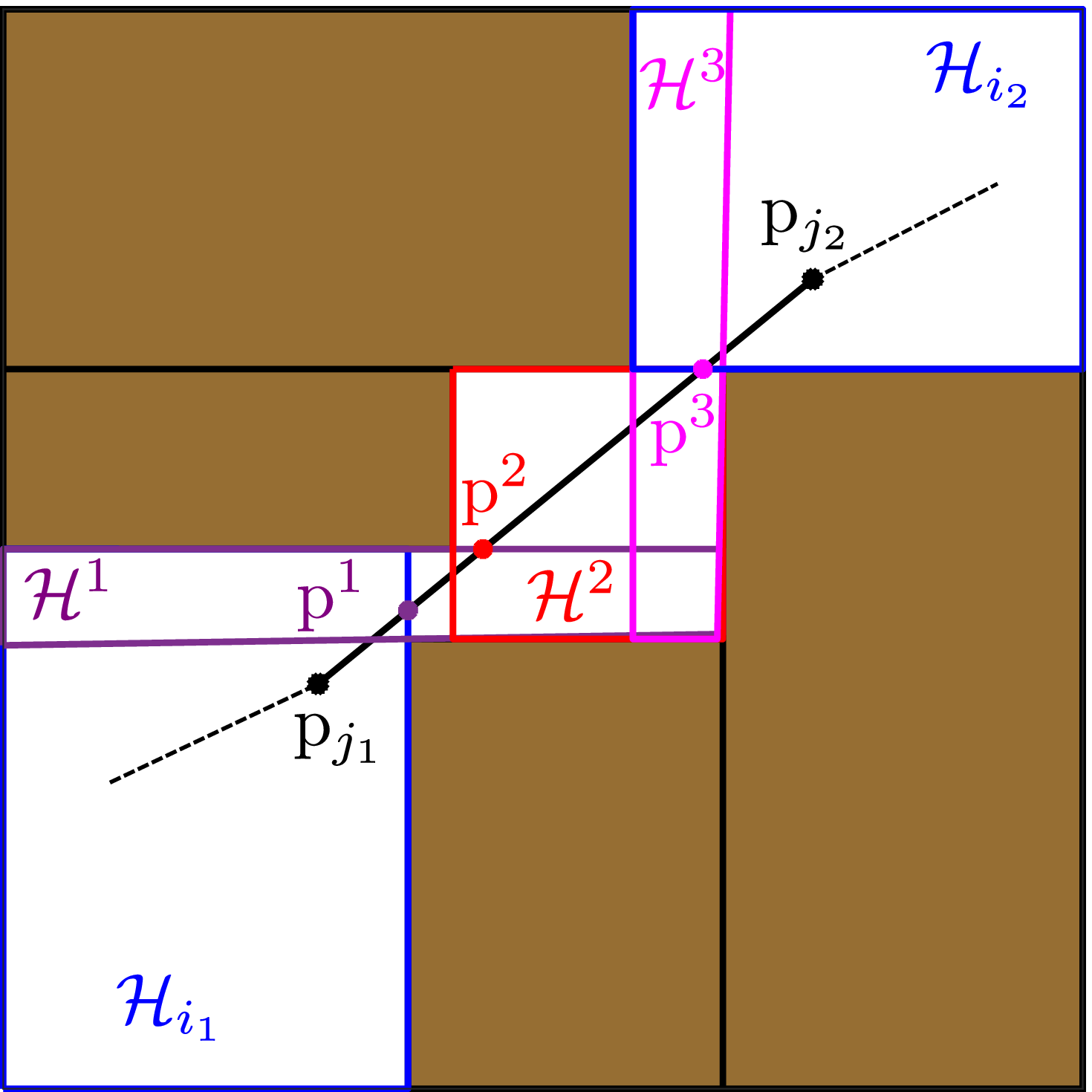}
\label{fig:disjoint_polytopes}
}
\centering
\subfigure[]
{
\includegraphics[width=0.22\textwidth]{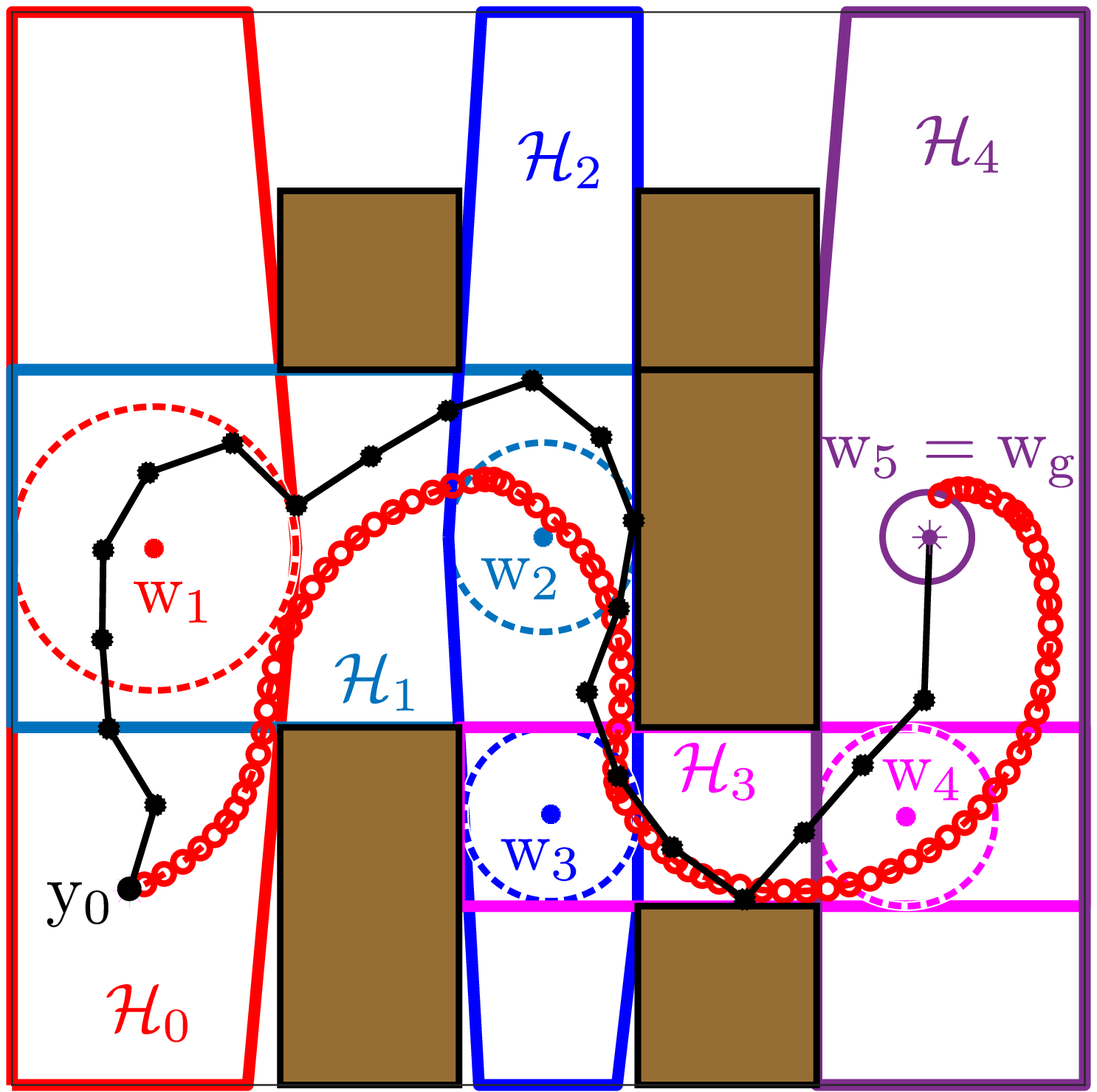}
\label{fig:inersecting_polytopes}
}
\vskip -5pt
\caption{(a) The workings of function $\mathsf{Intersecting\_Polytopes}$. The function takes a pair of non-intersecting polytopes $\mathcal{H}_{i_1}$ and $\mathcal{H}_{i_2}$ generated by the seed points $\mathrm{p}_{j_1}$ and $\mathrm{p}_{j_2}$, respectively, and computes the point $\mathrm{p^1}$ at which the line segment from $\mathrm{p}_{j_1}$ to $\mathrm{p}_{j_2}$ intersects $\mathcal{H}_{i_1}$. This point is then used as a new seed point to obtain a new polytope $\mathcal{H}^1$. Then, the point $\mathrm{p}^2$ at which the segment from $\mathrm{p^1}$ to $\mathrm{p}_{j_2}$ intersects the new polytope $\mathcal{H}^1$ is computed and used as a new seed to obtain $\mathcal{H}^2$. The process continues until $\mathcal{H}^{\ell}$ intersects $\mathcal{H}_{i_2}$. (b) Given $(\mathcal{H}_i, {\rm w}_{i+1})$, $i=0,...,4$ obtained by Algorithm~\ref{alg:RRT} based on an \ac{RRT} plan $\Pi$ (black), a sequence of MPCs is defined that outputs a collision-free path (red) connecting the initial $\mathrm{y}_0$ and goal $\mathrm{w}_\mathrm{g}$ locations.}
\label{fig:intersecting_polytopes}
\vskip -10pt
\end{figure}

\begin{itemize}
    \item $\mathsf{Generate\_Polytopes}$: Given the static map $\mathcal{W}^\mathrm{s}$ and a ``seed'' point $\mathrm{p}$ in the free space $\mathcal{W} \setminus \mathcal{W}^\mathrm{s}$, this function creates a polytope $\mathcal{H} \subset \mathcal{W} \setminus \mathcal{W}^\mathrm{s}$ . This is done in a convex optimization fashion via the algorithm in~\cite{Deits2015_IRIS}; here, the algorithm is terminated so that the seed point $\mathrm{p}$ is included in the interior of the returned polytope $\mathcal{H}$. 
    \item $\mathsf{Chebyshev\_Center}$: This function takes a pair of polytopes $\mathcal{H}_{i_1}$ and $\mathcal{H}_{i_2}$ and returns the Chebyshev center\footnote{The Chebyshev center of two intersecting polytopes is the center of the largest ball inscribed in their intersection~\cite{Boyd_CO}.} $\rm w$ of their intersection $\mathcal{H}_{i_1} \bigcap \mathcal{H}_{i_2}$, or an empty list if they are disjoint. Determining whether two polytopes intersect and computing the corresponding Chebyshev center is done via a linear program~\cite[Section 4.3]{Boyd_CO}.
    \item $\mathsf{Intersecting\_Polytopes}$: This function takes two seed points $\mathrm{p}_{j_1}$ and $\mathrm{p}_{j_2}$ from $\Pi$ together with the corresponding polytopes $\mathcal{H}_{i_1}$ and $\mathcal{H}_{i_2}$ generated by $\mathsf{Generate\_Polytopes}$, and returns a sequence of pairs 
    \begin{equation} \nonumber
       \bar{\mathcal{G}} = \{(\mathcal{H}^\ell, {\rm w}^\ell) ~|~ \ell = 1,..., L+1\} 
    \end{equation}
    where $\mathcal{H}^{L+1} = \mathcal{H}_{i_2}$. The chain $\{ \mathcal{H}_{i_1}, \mathcal{H}^1, ..., \mathcal{H}^L, \mathcal{H}_{i_2} \}$ is constructed as in Fig.~\ref{fig:intersecting_polytopes} so that it includes sequentially pair-wise intersecting polytopes, effectively connecting $\mathcal{H}_{i_1}$ and $\mathcal{H}_{i_2}$. The points ${\rm w}^1 \in \mathcal{H}_{i_1} \bigcap \mathcal{H}^1$ and ${\rm w}^\ell \in \mathcal{H}^{\ell-1} \bigcap \mathcal{H}^\ell$ for $\ell = 2,..., L+1$ are the corresponding Chebyshev centers. When $\mathcal{H}_{i_1}$ and $\mathcal{H}_{i_2}$ intersect, $L=0$ and the function returns $(\mathcal{H}^1, {\rm w}^1)$ where $\mathcal{H}^1 = \mathcal{H}_{i_2}$. 
    \item $\mathsf{Poly\_Line\_Intersect}$: This function takes a polytope $\mathcal{H}$, a point $\mathrm{p}_1 \in \mathcal{H}$ and a point $\mathrm{p}_2 \notin \mathcal{H}$ and returns the point at which the line segment connecting $\mathrm{p}_1$ and $\mathrm{p}_2$ intersects the boundary of $\mathcal{H}$. 
\end{itemize}

Algorithm~\ref{alg:RRT} begins with calling $\mathsf{Generate\_Polytopes}$ to create a polytope $\mathcal{H}_0$ in $\mathcal{W} \setminus \mathcal{W}^\mathrm{s}$ that contains the initial location $\mathrm{y}_0= C \mathrm{x}_0$. The algorithm then checks if the goal $ \mathrm{w}_\mathrm{g} \in \mathcal{H}_0$, and, if not, it runs through the points $\mathrm{p}_j$, $j=1,...,K$, of the plan $\Pi$ to find the first point $\mathrm{p}_{j_1}$ that is not contained in $\mathcal{H}_0$. The pairs $(\mathcal{H}_0, \mathrm{y}_0)$ and $(\mathcal{H}_\mathrm{new}, \mathrm{p}_{j_1})$ are then passed to $\mathsf{Intersecting\_Polytopes}$. If $\mathcal{H}_0 \bigcap \mathcal{H}_{\rm new} \neq \emptyset$, the function returns $(\mathcal{H}^1, {\rm w}^1)$ where $\mathcal{H}^1=\mathcal{H}_\mathrm{new}$ and ${\rm w}^1$ is the Chebyshev center of the intersection $\mathcal{H}_0 \bigcap \mathcal{H}_{\rm new}$. If, on the other hand, $\mathcal{H}_0 \bigcap \mathcal{H}_{\rm new} = \emptyset$, the function produces a sequence of pairs of polytopes and Chebyshev centers $\bar{\mathcal{G}}$ connecting $\mathcal{H}_0$ and $\mathcal{H}_{\rm new}$. The end result is a sequence of pairwise intersecting polytopes $\{ \mathcal{H}_0, \mathcal{H}_1, ..., \mathcal{H}_{M-1} \}$ with $\mathrm{y}_0 \in \mathcal{H}_0$ and waypoints $\{{\rm w}_1, ..., {\rm w}_M \}$ such that 
\begin{align}
    \label{eq:composition-2}
    {\rm w}_i &\in \mathcal{H}_{i-1} \bigcap \mathcal{H}_i ~~\text{for}~~ i=1,..., M-1 
\end{align}
and ${\rm w}_M = \mathrm{w}_{\rm g} \in \mathcal{H}_{M-1}$.

\begin{rem}\label{rem:rewire}
Algorithm~\ref{alg:RRT} is executed offline. In general, if the initial $\mathrm{y}_0$ and/or goal $\mathrm{w}_{\rm g}$ locations change, Algorithm~\ref{alg:RRT} must be executed anew. This is not necessary, however, when $\mathrm{y}_0$, $\mathrm{w}_{\rm g}$ are in the union of an available sequence of polytopes.

For example, if $\{\mathcal{H}_0,...,\mathcal{H}_{M-1}\}$ is available and $\mathrm{y}_0$, $\mathrm{w}_{\rm g}$ change so that $\mathrm{y}_0 \in \mathcal{H}_m$ and $\mathrm{w}_{\rm g} \in \mathcal{H}_n$ for some $m$ and $n$, then the given sequence can be ``rewired'' to a new sequence $\{ \mathcal{H}_m,\mathcal{H}_{m+1},...,\mathcal{H}_n \}$ if $m<n$ or $\{ \mathcal{H}_m,\mathcal{H}_{m-1},...,\mathcal{H}_n\}$ if $m>n$ without the need to execute Algorithm~\ref{alg:RRT} again.
\end{rem}

\subsection{Sequential Model Predictive Control}
\label{subsec:SeqMPC}

The directed decomposition of the free space by  Algorithm~\ref{alg:RRT} is used to define a sequence of \ac{MPC}s that steer the \ac{LIP} to the goal while avoiding static and dynamic obstacles. In what follows, we assume that an ordered collection of pairs $(\mathcal{H}_i, {\rm w}_{i+1})$, $i=0,...,M-1$, satisfying \eqref{eq:composition-2} is available.   

\subsubsection{Static obstacles}
Each polytope $\mathcal{H}_i$, $i=0,...,M-1$ is free from collisions with the static obstacles, and can be represented as a finite collection of $l_i$ closed half-spaces
\begin{equation} \nonumber
    \mathcal{H}_i = \{\mathrm{y} \in \mathbb{R}^2 ~|~ 
    P_i \mathrm{y} \leq b_i \}
\end{equation}
where $P_i$ is an $l_i \times 2$ matrix and $b_i \in \mathbb{R}^{l_i}$. As usual, $P_i \mathrm{y} \leq b_i$ is the shorthand notation for the system of inequalities $(r_i)^\mathsf{T}_j \mathrm{y} \leq (b_i)_j$, $j=1,...,l_i$, where $(r_i)^\mathsf{T}_j$ is $j$-th row of $P_i$ and $(b_i)_j$ the $j$-th element of $b_i$. To each $\mathcal{H}_i$ we can then associate $l_i$ smooth scalar-valued functions $h_{ij} : \mathbb{R}^5 \rightarrow \mathbb{R}$, 
\begin{equation}\label{eq:barrier_polytope}
    h_{ij}(\mathrm{x}) = (b_i)_j - (r_i)^\mathsf{T}_j C \mathrm{x}, ~~ j=1,...,l_i 
\end{equation}
so that the intersection of the corresponding 0-superlevel sets 
\begin{equation} \label{eq:safe_states}
    \mathcal{C}_{ij} = \{ \mathrm{x} \in \mathbb{R}^5 ~|~ h_{ij}(\mathrm{x}) \geq 0 \}, ~~ j=1,...,l_i    
\end{equation}
contains the states of the 3D-\ac{LIP} \eqref{eq:LIP_output} that are mapped via the output \eqref{eq:LIP_output} onto the free polytopes $\mathcal{H}_i$. Let $\mathcal{C}_i = \bigcap^{l_i}_{j = 1} \mathcal{C}_{ij}$ be the intersection of the (unbounded) polyhedra \eqref{eq:safe_states}.

With these definitions, collision-free evolution is achieved by selecting the input of the 3D-\ac{LIP} so that its state is constrained to evolve in the corresponding $\mathcal{C}_i$ when it starts in $\mathcal{C}_i$. 
If for some state $\mathrm{x}_0$ we have $h_{ij}(\mathrm{x}_0) \geq 0$ for all $j=1,...,l_i$ (that is, $\mathrm{x}_0 \in \mathcal{C}_i$), evolution in $\mathcal{C}_i$ can be achieved by choosing  $\mathrm{u}_k$ so that the following inequalities are simultaneously (over $j$) satisfied 
\begin{equation}\label{eq:CBF_constraint}
    h_{ij} \left( A \mathrm{x}_k + B \mathrm{u}_k \right) \geq (1-\gamma) h_{ij}\left(\mathrm{x}_{k}\right), ~~ j=1,...,l_i
\end{equation}
where $0 < \gamma \leq 1$. The existence of an input value $\mathrm{u}_{k}$ that satisfies \eqref{eq:CBF_constraint} implies that $h_{ij}$ is a discrete-time exponential \ac{CBF}~\cite{Agrawal2017RSS} for the 3D \ac{LIP} dynamics \eqref{eq:LIP}. Note that for each $i = 0,..., M-1$ and each $j = 1, ..., l_i$, the constraint \eqref{eq:CBF_constraint} is affine in $\mathrm{u}_k$ and will be incorporated in an \ac{MPC} program defined for each $i = 0,..., M-1$ to ensure the system is safe with respect to collisions with static obstacles.

\subsubsection{Moving obstacles}
Let $\mathcal{E}_{\nu,k} \subset \mathcal{W}$ be the area covered at time $k$ by the $\nu$-th moving obstacle, $\nu = 1, ..., n^\mathrm{d}$. We consider moving obstacles of elliptical shape, suitably inflated as in~\cite{mora2019RAL} to account for the robot's nontrivial dimensions. Thus, $\mathcal{E}_{\nu,k}$ can be represented by the pair $(\mathrm{p}^\mathrm{d}_{\nu,k}, P^\mathrm{d}_\nu(\varphi_{\nu,k}))$, where $\mathrm{p}^\mathrm{d}_{\nu,k}$ is the center of the $\nu$-th ellipse, $\varphi_{\nu,k}$ its orientation, and the matrix $P^\mathrm{d}_{\nu}(\varphi_{\nu,k})$ captures its shape~\cite{mora2019RAL}; see Appendix~\ref{app:moving-obstacles} for more details.
Assuming that $\mathrm{p}^\mathrm{d}_{\nu,k}$ and $\varphi_{\nu,k}$ are known at each $k$, collision with the $\nu$-th obstacle is avoided by imposing the barrier constraint  
\begin{equation}\label{eq:barrier_moving}
    h^{\rm d}_{\nu,k+1}(A \mathrm{x}_k + B \mathrm{u}_k)\geq (1-\gamma_\nu)h^{\rm d}_{\nu,k}(\mathrm{x}_{k}),~\nu = 1,..., n^\mathrm{d}
\end{equation}
where $0 < \gamma_{\nu} \leq 1$ and
\begin{equation} \nonumber
    h^{\rm d}_{\nu,k}(\mathrm{x}_k) = 
    \big(C \mathrm{x}_k - \mathrm{p}^\mathrm{d}_{\nu,k} \big)^{\! \mathsf{T}}  
    P^\mathrm{d}_\nu(\varphi_{\nu,k}) 
    \big(C\mathrm{x}_k - \mathrm{p}^\mathrm{d}_{\nu,k}\big) - 1 \enspace.
\end{equation}

\subsubsection{Sequential MPC}
Given the ordered collection $(\mathcal{H}_i, {\rm w}_{i+1})$, $i = 0,..., M-1$, of free polytopes and waypoints satisfying \eqref{eq:composition-2}, we define here a corresponding sequence MPC$(i)$, $i = 0,..., M-1$, of \ac{MPC} programs. In this sequence, the objective of the $i$-th \ac{MPC} is to drive the output \eqref{eq:LIP_output} of the 3D-\ac{LIP} towards the $(i+1)$ waypoint ${\rm w}_{i+1}$ while keeping it within $\mathcal{H}_i$, avoiding moving obstacles, and satisfying the relevant state and input constraints \eqref{eq:constraint_set}. Then, composing the MPCs according to the sequence 
\begin{equation}\nonumber
    \cdots \longrightarrow 
    \text{MPC}(i) \xrightarrow[]{C \mathrm{x}_k \in \mathcal{H}_{i+1}} \text{MPC}(i+1)
    \longrightarrow \cdots
\end{equation}
for $i = 0,...,M-1$ results in a sequence of control values that drives the 3D-\ac{LIP} to the goal location ${\rm w_g} = {\rm w}_M \in \mathcal{H}_M$. Switching from MPC$(i)$ to MPC$(i+1)$ is triggered when the output $C \mathrm{x}_k$ enters the polytope $\mathcal{H}_{i+1}$.

The objective of MPC$(i)$ is captured by the desired state
\begin{equation}\label{eq:MPC_target}
    S\mathrm{x}^{\rm des}_{i+1} = 
    \begin{bmatrix} {\rm w}^\mathsf{T}_{i+1} & \theta_{i+1} & 0_{1 \times 2} 
    \end{bmatrix}^\mathsf{T}
\end{equation}
where $S$ is the $5 \times 5$ selection matrix that re-organizes the state of the 3D-\ac{LIP} so that $S\mathrm{x} = \begin{bmatrix} x & y & \theta & \dot{x} & \dot{y} \end{bmatrix}^\mathsf{T}$, and ${\rm w}_{i+1}$ is the target waypoint. The desired orientation $\theta_{i+1}$ is set to be the angle of the vector that connects the position $C \mathrm{x}_{k}$ of the 3D-\ac{LIP} at step $k$ with ${\rm w}_{i+1}$. 

The terminal cost of MPC$(i)$ can now be defined by 
\begin{equation}\label{eq:MPC_terminal}
    J_{i,N}(\mathrm{x}_N) = \| S\mathrm{x}_N - S\mathrm{x}^{\rm des}_{i+1}\|^2_{W_1}
\end{equation}
where $W_1$ is a $5 \times 5$ diagonal weighting matrix. The running cost is constructed similarly to \eqref{eq:MPC_terminal} with the modification of including an additional term that penalizes the magnitude of the 3D-\ac{LIP} input at each step; that is, 
\begin{equation}\label{eq:MPC_running}
    J_{i,k}(\mathrm{x}_k,\mathrm{u}_k) = \| S\mathrm{x}_k - S\mathrm{x}^{\rm des}_{i+1}\|^2_{W_2} + \|\mathrm{u}_k\|^2_{W_3}
\end{equation}
where $W_2$ and $W_3$ are $5 \times 5$ diagonal weighting matrices. Note here that, by \eqref{eq:MPC_target}, minimizing  \eqref{eq:MPC_terminal}-\eqref{eq:MPC_running} also penalizes the velocity of the \ac{LIP}, effectively introducing a damping effect which is necessary since the barrier polyhedra $\mathcal{C}_i$ do not bound the \ac{LIP} velocity. This effect facilitates the execution of the \ac{LIP}-based plan by Digit. 

To summarize, the $i$-th MPC is defined as:
\noindent\rule{\columnwidth}{0.5pt}

\textbf{MPC$(i)$:}
\begin{subequations}
\begin{IEEEeqnarray}{s'rCl'rCl'rCl}
\nonumber 
$\underset{X, U}{\text{minimize}}$ & \IEEEeqnarraymulticol{9}{l} {\displaystyle \sum^{k+N-1}_{\kappa=k} J_{i,\kappa}(\mathrm{x}_{\kappa}, \mathrm{u}_\kappa) + J_{i,N}(\mathrm{x}_N)} \\
\nonumber
subject to & \mathrm{x}_{\kappa+1} &=& A\mathrm{x}_{\kappa} + B\mathrm{u}_{\kappa} \\
\nonumber
& \mathrm{x}_k &=& \mathrm{x}_\mathrm{init}, ~(\mathrm{x}_\kappa, \mathrm{u}_\kappa) \in \mathcal{XU}_\kappa \\ 
\nonumber
& \Delta h_{i j}(\mathrm{x}_\kappa, \mathrm{u}_\kappa) & \geq & -\gamma h_{i j}(\mathrm{x}_\kappa), ~j = 1,...,l_i \\
\nonumber
& \Delta h^{\rm d}_{\nu,\kappa}(\mathrm{x}_\kappa, \mathrm{u}_\kappa) & \geq & -\gamma_\nu h^{\rm d}_{\nu,\kappa}(\mathrm{x}_\kappa), ~\nu = 1,...,n^\mathrm{d}
\end{IEEEeqnarray}
\end{subequations}
\noindent\rule{\columnwidth}{0.5pt}
where $\mathcal{XU}_\kappa$ is the constraint set \eqref{eq:constraint_set}, $h_{i j}$ and $h^{\rm d}_{\nu,\kappa}$ are the barrier functions \eqref{eq:barrier_polytope} and \eqref{eq:barrier_moving}, and $\Delta h_{i j}(\mathrm{x}_\kappa, \mathrm{u}_\kappa) = h_{i j}(\mathrm{x}_{\kappa+1}) - h_{i j}(\mathrm{x}_\kappa)$ and $\Delta h^{\rm d}_{\nu,\kappa}(\mathrm{x}_{\kappa}, \mathrm{u}_\kappa) = h^{\rm d}_{\nu,{\kappa+1}}(\mathrm{x}_{\kappa+1})-h^{\rm d}_{\nu,\kappa}(\mathrm{x}_\kappa)$. 
Given $\mathrm{x}_\mathrm{init}$, the solution of MPC$(i)$ returns the sequences $X = \{ \mathrm{x}_{k+1},...,\mathrm{x}_{k+N} \}$ and $U = \{\mathrm{u}_k,...,\mathrm{u}_{k+N-1} \}$. We then apply the first element $\mathrm{u}_k$ in $U$ and proceed with solving MPC$(i)$ at the next step  based on the new initial condition $\mathrm{x}_{\rm init}$. The process is repeated  until $C \mathrm{x}_k \in \mathcal{H}_{i+1}$ for some $k$, where we switch to MPC$(i+1)$ and continue until the goal.

\subsection{Numerical Experiments and Performance Evaluation}
\label{sec:MPC_Perf}

\begin{figure}[b]
\centering
\subfigure[]{\includegraphics[width=0.23\textwidth]{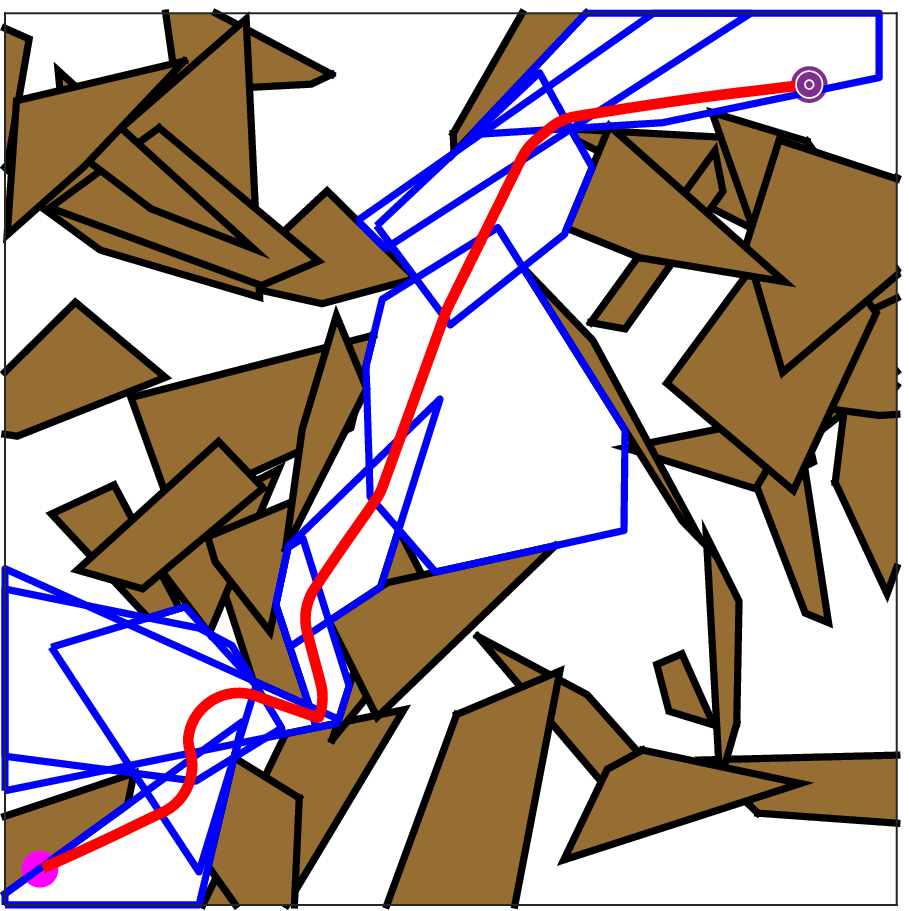}}\quad
\subfigure[]{\includegraphics[width=0.23\textwidth]{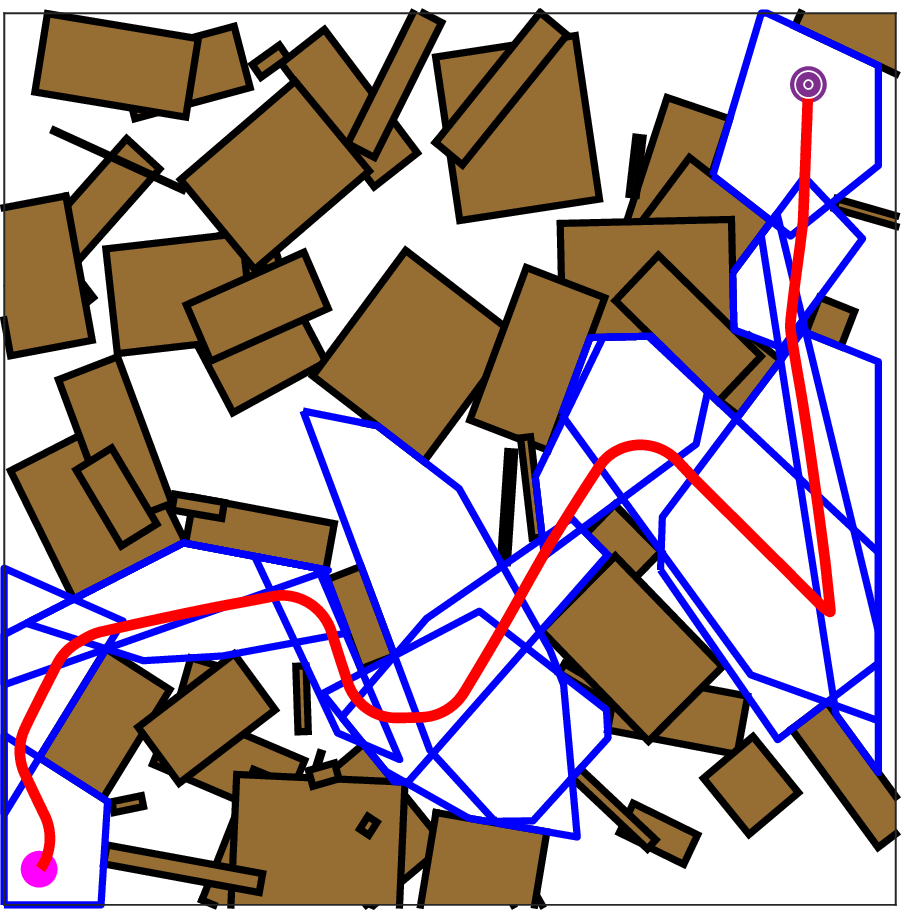}}\quad
\vskip -5pt
\caption{Example environments and paths generated by the \ac{MPC}. (a) Polygonal obstacles. (b) Rotated rectangular obstacles.}
\label{fig:cluttered_obstacles}
\end{figure}

In this section, we numerically investigate the effectiveness of our approach in terms of its ability to generate collision-free paths and the associated computation time. The quality of realizing the planned path on Digit will be considered in the following section. We consider here three sets of randomly generated environments containing
\begin{inparaenum}[(i)]
\item rectangular obstacles with different sizes,
\item rectangular obstacles with different sizes and orientations,
\item general polytopic obstacles;
\end{inparaenum}
Fig.~\ref{fig:cluttered_obstacles} shows typical instances. In each case, we generate $50$ random environments with $30, 40, 50$ and $60$ obstacles in a confined $50 \rm{m} \times 50 {\rm m}$ space; hence, $150$ environments are considered for each obstacle population, giving a total of $600$ static maps. The obstacle coverage is roughly 40\% in all maps, and the starting and goal locations are kept the same. We employ Algorithm~\ref{alg:RRT} to generate sequentially intersecting free polytopes and waypoints that provide constraints and objectives for the \ac{MPC}s. The \ac{LIP} parameters and are given in Appendix~\ref{app:numerical-experiments}; Table~\ref{table:1} summarizes our results.

We note first that, given the \ac{RRT} path, Algorithm~\ref{alg:RRT} was able to compute a sequence of obstacle-free polytopes in all the environments considered. The average computation time required for the polytopic decomposition per map is shown in Table~\ref{table:1}, and it increases from $226 \text{ms}$ for $30$ obstacles to $343 \text{ms}$ when the number of obstacles is doubled. Although Algorithm~\ref{alg:RRT} is typically executed offline, a new polytopic decomposition can be performed in less than a second, even for highly cluttered environments; recall that by Remark~\ref{rem:rewire} a new decomposition may not always be needed.

\begin{table}[h]
\centering
\caption{Average Computation time for Polytope Generation and \ac{MPC}}
\vskip -5pt
\begin{tabular}{|c | c | c | c | c|} 
 \hline
 \textbf{No. of Obstacles} & 30 & 40 & 50 & 60 \\ 
 \hline
 \textbf{Polytope Generation (ms)} & 226 & 274 & 300 & 343 \\
 \hline
 \textbf{MPC horizon $N = 2$ (ms)} & 16.8 & 16.5 & 16.3 & 16.7 \\
 \hline
 \textbf{MPC horizon $N = 3$ (ms)} & 23.3 & 23.7 & 23.4 & 23.8 \\
 \hline
 \textbf{MPC horizon $N = 4$ (ms)} & 39.1 & 39.1 & 39.4 & 39.6 \\
 \hline
\end{tabular}
\label{table:1}
\vskip -5pt
\end{table}

Table~\ref{table:1} also shows the average time required for solving the \ac{MPC} for $N = 2, 3, 4$. It can be seen that, for a given horizon, this time does not vary significantly with the number of static obstacles. This is because the \ac{MPC} does not consider the full environment at once; rather, it focuses on the part of the free space in the vicinity of the robot, capturing it in a polytope described by a small number of constraints. In other words, the \ac{MPC} does not take into account all the obstacles  simultaneously---as is the case in~\cite{Agrawal2017RSS, Ghaffari2021Arxiv} for example, where one \ac{CBF} per obstacle is included in the \ac{MPC} program---but it ``consumes'' the free space in a ``piecemeal'' fashion, as the system progresses towards the goal. This is beneficial, particularly in highly cluttered spaces that may contain a large number of obstacles.

Finally, as expected, the time required to solve the \ac{MPC} increases with the length $N$ of the horizon. For $N = 2$, the optimization is faster, but the \ac{MPC} failed to provide a solution in $8$\% of the environments considered. On the other hand, using $N = 3$ and $N = 4$ produced a path for all cases. Thus, in what follows, we will use $N = 3$ as a reasonable trade-off between feasibility and computational time.

\section{Realizing 3D-LIP Plans on Digit}
\label{sec: Digit}

We apply the proposed approach to planning motions for the bipedal robot Digit (cf. Fig.~\ref{fig:DigitStructureandFoot}) walking amidst static and moving obstacles. Digit was designed and manufactured by Agility Robotics\footnote{https://www.agilityrobotics.com/robots\#digit}; it weighs $47.9 \text{kg}$ and it is approximately $1.58 \text{m}$ tall when standing up. It shares the bird-like leg design of its predecessor, Cassie~\cite{gong2019ACC}, albeit Digit's legs terminate at non-trivial feet capable of realizing planar contacts with the ground. Each leg features 8 \ac{DOF}, two of which correspond to passive leaf springs (cf.  Fig.~\ref{fig:DigitStructureandFoot}). Furthermore, Digit incorporates a pair of 4 \ac{DOF} arms that extend both the mobility and utility of the robot. Thus, the robot has a total of 30 \ac{DOF}s, 20 of which are directly actuated, 4 are passive and 6 correspond to the position and orientation of a body-fixed frame (cf. Fig.~\ref{fig:DigitStructureandFoot}).

\vspace{-0.05in}
\subsection{Digit Dynamics: Notation and Assumptions}

\begin{figure}[t]
\centering
{
\includegraphics[width=0.99\columnwidth]{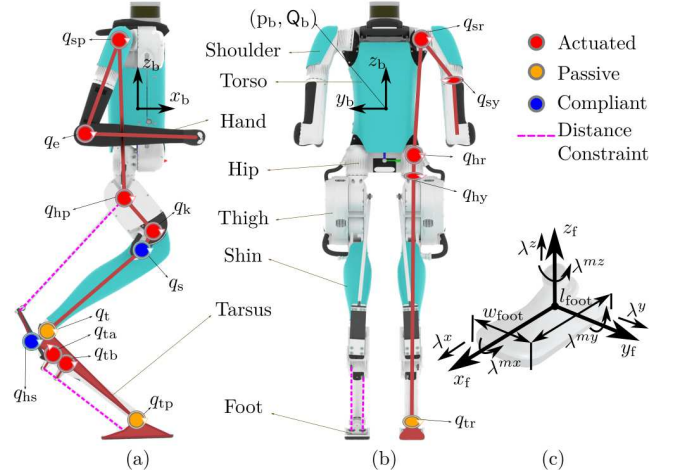}
}
\vskip -15pt
\caption{The bipedal robot Digit and relevant coordinates. (a) Side view; the kinematic loops actuating the tarsus and foot are highlighted with the dotted magenta lines. (b) Back view. (c) Ground contact constraints.}
\label{fig:DigitStructureandFoot}
\end{figure}

Digit's model is a tree structure of interconnected rigid bodies representing the robot’s torso, legs and arms. Its configuration can be described with respect to an inertia frame by the position $\mathrm{p}_\mathrm{b} \in \mathbb{R}^3$ and the unit quaternion $\mathsf{Q}_\mathrm{b} \in \mathsf{SO}(3)$ associated with the orientation of a body-fixed frame $\{x_\mathrm{b}, y_\mathrm{b}, z_\mathrm{b}\}$ attached at Digit's torso (cf. Fig.~\ref{fig:DigitStructureandFoot}). Each leg $\iota \in \{\mathrm{L}, \mathrm{R}\}$ is connected with the torso at the hip, which is parametrized by the corresponding yaw, roll and pitch angles $(\Qi{hy}, \Qi{hr}, \Qi{hp})$. The legs are composed by the thigh, shin and tarsus links interconnected via  compliant four-bar linkages captured by the knee $\Qi{k}$, the shin-spring $\Qi{ss}$, the tarsus $\Qi{t}$, and the heel-spring $\Qi{hs}$ angles, as shown in Fig.~\ref{fig:DigitStructureandFoot}. The legs terminate at feet, the orientation of which is described by the corresponding roll $\Qi{tr}$ and pitch $\Qi{tp}$ angles; the feet are actuated via rigid four-bar linkages parametrized by the angles $\Qi{ta}$ and  $\Qi{tb}$. Finally, each arm $\iota \in \{\mathrm{L}, \mathrm{R}\}$ is parametrized by the yaw, roll and pitch angles $(\Qi{sy}, \Qi{sr}, \Qi{sp})$ at the shoulder joint, and by the angle $\Qi{e}$ of the elbow joint. Overall, each leg/arm pair features 15 coordinates 
\begin{align}\nonumber
    q^{\iota \in\{\mathrm{L,R}\}} = \big( &\Qi{hy},~\Qi{hr},~\Qi{hy},~\Qi{k},~\Qi{ss},
    ~\Qi{t}, ~\Qi{hs},\\ 
    \nonumber 
    &\Qi{ta},~\Qi{tb},~\Qi{tr},~\Qi{tp},
    \Qi{sr},\Qi{sp},\Qi{sy},\Qi{e} \big) \in \mathcal{Q}^{\iota \in\{\mathrm{L,R}\}}
\end{align}
and the configuration of the free-floating model is  
\begin{equation} \nonumber
    q= \left( \mathrm{p}_\mathrm{b},\mathsf{Q}_\mathrm{b}, q^\mathrm{L}, q^\mathrm{R} \right) \in \mathbb{R}^3 \times \mathsf{SO}(3) \times \mathcal{Q}^\mathrm{L}_\mathrm{s} \times \mathcal{Q}^\mathrm{R}_\mathrm{s} \enspace.  
\end{equation}
  
Walking is composed of alternating sequences of single and double support phases. For our controller design, we will focus on the single support phase and assume that the deflections of the leaf springs are negligible. Under these assumptions, the single-support dynamics of Digit are    
\begin{equation}\label{eq: eom_constrained}
    \begin{bmatrix}
    D(q) & J(q)^\mathsf{T}\\ J(q) & 0
    \end{bmatrix}
    \begin{bmatrix}
    \phantom{-}\dot{v}\\ -\lambda
    \end{bmatrix}
    =
    \begin{bmatrix}
    -c(q,v) + S^\mathsf{T}_\tau \tau\\
    -\dot{J}(q,v)v
    \end{bmatrix}
\end{equation}
where $D(q)$ is the inertia matrix, $c(q,v)$ contains the velocity-dependent and gravitational forces, $S_\tau$ is the input selection matrix and ${\tau}\in\mathbb{R}^{20}$ are the motor torques. Due to space limitations, we do not provide a detailed account of how the kinematics loops are treated; this is done in a similar fashion as in Cassie~\cite{Reher2021ICRA}. We only mention here that the Jacobian $J(q)$ incorporates all the  holonomic constraints associated with the single-support phase and $\lambda\in\mathbb{R}^{16}$ are the corresponding constraint generalized forces. Finally, contact limitations are satisfied by requiring the contact wrench $(\lambda^x, \lambda^y, \lambda^z, \lambda^{mx}, \lambda^{my}, \lambda^{mz})$ to lie in a linearized \ac{CWC} $\mathcal{K}$, as in~\cite[Proposition 2]{Caron2015ICRA}.

\vspace{-0.05in}
\subsection{Digit Control: Weighted Operational Space QP}
\label{subsec:Digit_control}

An \ac{OSC} formulated as a weighted \ac{QP} is used to command suitable inputs to Digit's actuators so that the following locomotion objectives are achieved (cf. Appendix~\ref{app:OSC-QP} for more details).

\subsubsection{3D-LIP following}
The first objective is twofold; first, realize the suggested foot placement $\mathrm{u}_k$ obtained based on the most recent solution of the \ac{MPC}, and second, impose the trajectory of the 3D-\ac{LIP} on Digit's \ac{COM}. Consider step $(k-1)$ and let  $s=(t-t_{k-1})/T$, $t\in[t_{k-1},t_k)$ be the phase variable.

The \ac{MPC} is solved within the $k-1$ step by using feedback from Digit to predict the state $\mathrm{x}_\mathrm{init}$ of the \ac{LIP} at the beginning of the next step.  
Given the result $\mathrm{u}_k = \begin{bmatrix} u^x_k & u^y_k & u^\theta_k \end{bmatrix}^\mathsf{T}$ of the \ac{MPC}, we define the desired trajectory for the position of the swing foot as in \cite{Gong2021Arxiv}, i.e.
\begin{equation}\nonumber
    \mathrm{p}_\mathrm{sw}^\mathrm{des}(s)=
    \begin{bmatrix}
    \mathrm{p}_{\mathrm{sw}}^x(t_{k-1}) + \frac{1}{2} \left( 1-\cos \pi s \right) u^x_k \\ 
    \mathrm{p}_{\mathrm{sw}}^y(t_{k-1}) + \frac{1}{2} \left( 1-\cos \pi s \right) u^y_k \\
    z_\mathrm{cl} \left( 1-4(s-0.5)^2 \right)
    \end{bmatrix}
\end{equation}
where $\mathrm{p}_\mathrm{sw}^x(t_{k-1})$ and $\mathrm{p}_\mathrm{sw}^y(t_{k-1})$ are the $x$ and $y$ coordinates of the swing foot at the beginning of step $(k-1)$ and $z_\mathrm{cl}$ is a desired clearance height at the middle of the step.
For the orientation of the swing foot, the desired trajectory in quaternion coordinates $\mathsf{Q}_\mathrm{sw}^\mathrm{des}(s)$ is defined by interpolating between the initial\footnote{Notation: $\mathsf{Q}(\phi^{x},\phi^{y},\phi^{z})$ is the quaternion obtained from the corresponding roll-pitch-yaw angles $\phi^{x}$, $\phi^{y}$, and $\phi^{z}$.} $\mathsf{Q}_\mathrm{sw}(t_{k-1}) = \mathsf{Q}(0, 0, \phi^z_\mathrm{sw}(t_{k-1}))$ and final $\mathsf{Q}_\mathrm{st}(t_k) = \mathsf{Q}(0, 0, \phi^z_\mathrm{sw}(t_{k-1}) + u^\theta_k)$ orientations of the foot by performing a spherical linear interpolation~\cite{Shoemake1985AnimatingRW} between the corresponding quaternions ($\mathsf{Slerp}$).

Next, we define the desired trajectory for Digit's torso based on updated predictions of the continuous-time trajectory of the \ac{LIP}. At time $t_{k-1}$ we predict the LIP trajectory $x(sT)$ by setting $x(0) = \mathrm{p}^x_\mathrm{COM}(t_{k-1})$, $\dot{x}(0) = \dot{\mathrm{p}}^x_\mathrm{COM}(t_{k-1})$ and $u^x = \mathrm{p}^x_\mathrm{COM}(t_{k-1}) - \mathrm{p}^x_\mathrm{st}(t_{k-1})$ in \eqref{eq:LIP_x_continuous}. Similarly, we predict the evolution $y(sT)$. Thus, the desired trajectory for the torso's \ac{COM} is defined as $\mathrm{p}^\mathrm{des}_\mathrm{COM}(s) = \begin{bmatrix} x(sT) & y(sT) & H \end{bmatrix}^\mathsf{T}$, where $H$ is the constant \ac{LIP} height. Note that the \ac{LIP} trajectory prediction can be updated within the current step $[t_{k-1}, t_{k})$, each time the \ac{MPC} is solved; see Appendix~\ref{app:Digit-MPC} for more details. This way, more recent information from Digit is incorporated, resulting in improved robustness. Finally, to maintain Digit's torso upright and pointing towards the direction of the motion, we define the desired orientation as $\mathsf{Q}_\mathrm{b}^{\mathrm{des}} = \mathsf{Q}(0,0,(\phi^{z}_\mathrm{st}+\phi^{z}_\mathrm{sw})/2)$, where $\phi^{z}_\mathrm{st}$ and $\phi^{z}_\mathrm{sw}$ are the yaw angles of the support and the swing feet, respectively.

Then, if $\eta_1 \in \mathbb{R}^{12}$ denotes the error from the desired trajectory, we can define the performance index associated with following the \ac{LIP}-based trajectory as 
 \begin{equation} \nonumber
     \Psi_1 = \left\| \ddot{\eta}_1 + K_\mathrm{D} \dot{\eta}_1 + K_\mathrm{P} \eta_1 \right\|^2
 \end{equation}
where $K_\mathrm{P}$ and $K_{\mathrm{D}}$ are suitably selected gains.

\subsubsection{Angular momentum}
Due to its point mass, the \ac{LIP} cannot capture the effect of the angular momentum on Digit. To mitigate this effect, we compute the angular momentum about Digit's \ac{COM} $\eta_2 = A_\mathrm{G}^\mathrm{ang} v$, where $A_\text{G}^{\text{ang}}$ is the angular part of the centroidal momentum matrix, and minimize
\begin{equation}\nonumber
     \Psi_2 = \left\|\dot{\eta}_2+K_{\mathrm{G}}\eta_2\right\|^2
\end{equation}
where $K_\mathrm{G}$ is a gain matrix.

\subsubsection{Arms desired configuration}
Given that $\Psi_1$ does not impose any restriction on the robot's arms, minimizing the angular momentum according to $\Psi_2$ may result in undesirable arm movements. To avoid this and maintain a nominal configuration for the arms, we define the following cost
\begin{equation} \nonumber
    \Psi_3 = \left\| \ddot{\eta}_3 + K_{\mathrm{D,a}}\dot{\eta}_3 + K_{\mathrm{ P,a}}(\eta_3-\eta^{\mathrm{des}}_3) \right\|^2
\end{equation}
where $\eta_3$ is the configuration of the left and right arms, $\eta^{\mathrm{des}}_3$ its desired value, and $K_{\mathrm{P,a}}$, $K_{\mathrm{D,a}}$ are suitable gains.  

\subsubsection{Input effort}
We penalize actuator effort by minimizing
\begin{equation} \nonumber
    \Psi_4 = \left\|\tau\right\|^2 \enspace.
\end{equation}
Introducing this term improves the performance of the controller by attenuating occasional spikes in the control signal.

\subsubsection{Weighted \ac{OSC}-\ac{QP}}
As in~\cite{mordatch2010ACM, Apgar2018RSS}, the aforementioned performance indices are combined in a single cost function with weights that reflect their respective level of importance, resulting in the following \ac{QP}:
\rule{\columnwidth}{0.4pt}
\textbf{OSC-QP:}
\begin{IEEEeqnarray*}{lClr}
\underset{\dot{{v}},\tau,\lambda}{\text{minimize}} &\quad& \psi_1 \Psi_1+\psi_2 \Psi_2+\psi_3 \Psi_3+\psi_4 \Psi_4  & \\
\text{subject to} 
    &&(\ref{eq: eom_constrained} )&(\text{Dynamics})\\
    &&lb_\tau\leq {\tau} \leq ub_\tau &(\text{Torque Limits})\\
    &&(\lambda^x, \lambda^y, \lambda^z, \lambda^{mx}, \lambda^{my}, \lambda^{mz})\in \mathcal{K} &(\text{\ac{CWC}})
\end{IEEEeqnarray*}
\rule{\columnwidth}{0.4pt}
In the cost, higher priorities are assigned larger weights. This QP is solved given $(q,v)$ at every step of the control loop.

\section{Simulation Results}
\label{sec:results}

We implemented the proposed sequential \ac{MPC} and \ac{OSC}-\ac{QP} controller to plan and execute obstacle-free motions in high-fidelity simulations with Digit. We use MuJoCo~\cite{Todorov2012IROS_mujoco} as our simulation environment with simulation loop running at $2 \mathrm{kHz}$. The dynamics quantities are calculated using spatial algebra~\cite{RBDA}, and used as symbolic functions generated with CasADi~\cite{Andersson2019_CasADi}.
An ordered collection $(\mathcal{H}_i, \mathrm{w}_{i+1})$, $i=1,..., M-1$ of free polytopes and waypoints is obtained by Algorithm~\ref{alg:RRT}, and the corresponding \ac{MPC}s are solved using IPOPT~\cite{Wchter2006IPOPT} solver shipped with CasADi~\cite{Andersson2019_CasADi}. The parameters for the \ac{MPC}s are given in Appendix~\ref{app:numerical-experiments}, and are the same as those used in Table~\ref{table:1}. However, in the implementation here, a solution for each \ac{MPC} is obtained in $\approx 10$ms, which much faster compared to the $\mathsf{fmincon}$ results of Table~\ref{table:1}. This way, $\mathrm{u}_k$ can be updated frequently in the light of more recent information from Digit, greatly improving the robustness of the planner and drastically reducing the error between the \ac{LIP}-suggested path and the one followed by Digit. 
The parameters for the \ac{OSC}-\ac{QP} implementation are given in Appendix~\ref{app:OSC-QP}. The \ac{QP} is solved using OSQP~\cite{boyd2020MPC} solver shipped with CVXPY~\cite{diamond2016cvxpy, agrawal2018JCD_rewriting} at $400 \mathrm{Hz}$. 
We limit the frequency of MPC to $15\mathrm{Hz}$ owing to the good tracking performance of QP.

\begin{figure*}
    \centering
    \includegraphics[width=\textwidth]{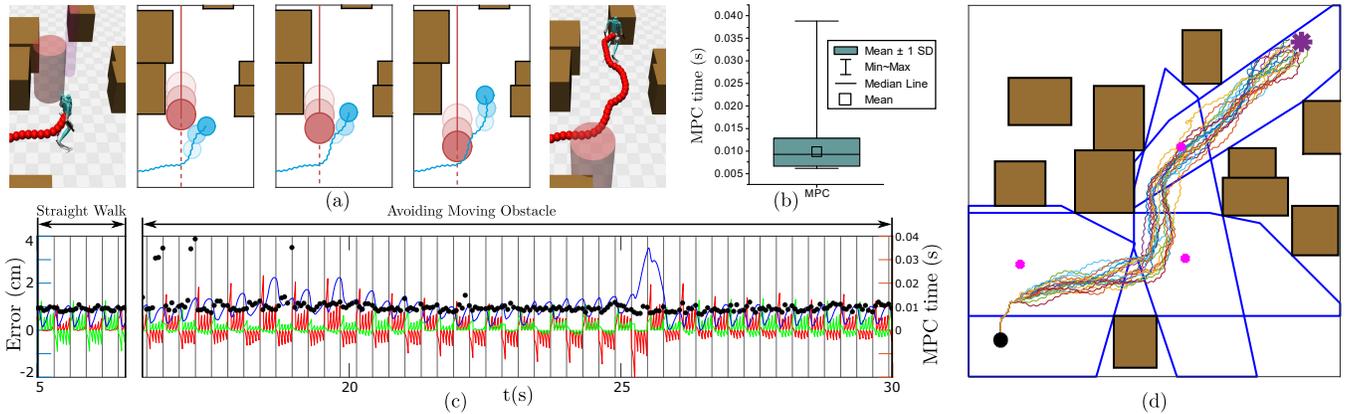}
    \vspace{-0.15in}
    \caption{(a) Snapshots of Digit (blue circle) traversing an obstacle course while avoiding an obstacle moving at $0.3 \mathrm{m/s}$ (red circle); for clarity, the middle tiles present the top view. (b) The average time required to solve the \ac{MPC} for the scenario in (a) is $\approx 10 \mathrm{ms}$ with occasional spikes at $\approx 40 \mathrm{ms}$ occurring at the early stages of encountering the moving obstacle. (c) Tracking error between the \ac{MPC} plan and Digit's \ac{COM}; errors in the $x$, $y$ and $z$ axes are in red, green, and blue, respectively. Black dots show the time required to solve the \ac{MPC}s. (d) Digit's \ac{COM} paths under randomly applied 3D forces.}
    \label{fig:Obstacle_avoidance}
\end{figure*}

We tested our method in different scenarios; the video accompanying this paper shows some instances. Here, we consider a $(10\mathrm{m} \times 10\mathrm{m})$ map with $10$ static obstacles. The environment contains a circular moving obstacle of radius $0.5 \mathrm{m}$ the motion of which is assumed to be known (cf. Fig.~\ref{fig:Obstacle_avoidance}(a)). The moving obstacle adds a constraint in the \ac{MPC}, which is taken into account when the obstacle is less than $5 \mathrm{m}$ away from the robot. This slightly increases the average time for solving the \ac{MPC} to $\approx 12 \mathrm{ms}$. We observe that the error between desired \ac{COM} trajectory of \ac{LIP} and the Digit remains small (cf. Fig.~\ref{fig:Obstacle_avoidance}(c)); the error increases when turning is required to avoid the moving obstacle, but never exceeds $3.5 \mathrm{cm}$. This implies that Digit's \ac{COM} faithfully executes the suggested \ac{LIP}-based plan.
Additionally, we tested the robustness of the method to  unexpected external forces applied at Digit's torso. The $x$, $y$ and $z$ components of the force were sampled uniformly over $[-50,50]~\mathrm{N}$ and the force was  applied for $100 \mathrm{ms}$ at random time instants separated at most $2 \mathrm{s}$. We conducted 30 trials, one out of which resulted in the robot failing to complete the task; Fig.~\ref{fig:Obstacle_avoidance}(d) shows the remaining 29 successful trials. Owing to the ability to solve the \ac{MPC}s at sufficiently high frequency, the robot was able to maintain balance and complete the objective of reaching the goal without collisions. The video~\cite{Video} presents various scenarios of Digit avoiding static and moving obstacles.

 \section{Conclusion}
%
We have presented a sequential \ac{MPC} framework for bipedal robots navigating through complex dynamic environments. The main idea is to decompose the free space as a sequence of mutually intersecting polytopes, each capturing locally the free space. This decomposition is then used to define a sequence of \ac{MPC} programs, within which collision avoidance is certified via discrete-time \ac{CBF}s. Finally, we tested our sequential \ac{MPC} framework in high-fidelity simulations with the bipedal robot Digit, demonstrating robust reactive obstacle avoidance in highly cluttered environments containing both static and moving obstacles.

\bibliographystyle{IEEEtran}
\bibliography{NaKouThaPo2022RAL_arxiv}

\begin{appendices}

\section{Details on LIP Constraints}
\label{app:LIP-constraints}

\subsubsection{Reachability and heading constraints}
The reachable regions for foot placement can be represented as
\begin{equation}\nonumber
    \begin{bmatrix}
    lb^{x_c}_k \\ lb^{y_c}_k
    \end{bmatrix}
    \!\leq\!
    \begin{bmatrix}
    \cos(\theta_k+u^{\theta}_k)            & -\sin(\theta_k+u^{\theta}_k) \\
    \sin(\theta_k+u^{\theta}_k) & \phantom{-}\cos(\theta_k+u^{\theta}_k)
    \end{bmatrix}^\mathsf{T}
    \!\!
    \begin{bmatrix}
    u^{x}_k \\  u^{y}_k 
    \end{bmatrix}
    \!\leq\!
    \begin{bmatrix}
    ub^{x_c}_k\\ub^{y_c}_k
    \end{bmatrix}
\end{equation}
where $lb$ and $ub$ denote the lower and upper bounds of the distance that can be reached along the forward $x_c$ and lateral $y_c$ directions. The bounds depend on $k$ because the robot switches its stance leg at each step; see Fig.~\ref{fig:3Dlip}(a).
We will limit the orientation update $u^\theta_k$ according to
\begin{equation}\nonumber
    lb^\theta \leq u^\theta_k \leq ub^\theta
\end{equation}
where $lb^\theta$ and $ub^\theta$ are suitable lower and upper bounds.

\subsubsection{COM travel constraints}
To avoid infeasible motions and bound the average speed of the COM between steps, we bound on the COM's travel during the step as follows,
\begin{equation}\nonumber
\label{eq:state_constraints}
    \delta_\mathrm{min}\leq\sqrt{\Delta x_k^2 + \Delta y_k^2}\leq \delta_\mathrm{max}
\end{equation}
where $\Delta x_k = x_{k+1}-x_k$ and $\Delta y_k = y_{k+1}-y_k$.

\section{Details on Moving Obstacles}
\label{app:moving-obstacles}

%
The shape matrix of $\nu$-th moving obstacle is given by
\begin{equation}\nonumber
    P^\mathrm{d}_{\nu}(\varphi_{\nu,k}) = R^{\rm d}(\varphi_{\nu,k})^\mathsf{T} 
    \begin{bmatrix} 
    1/\alpha^2_\nu & 0 \\ 0 & 1/\beta^2_\nu 
    \end{bmatrix}
    R^{\rm d}(\varphi_{\nu,k})
\end{equation}
where $\alpha_\nu$ and $\beta_\nu$ are the corresponding semi-axes suitably inflated as in~\cite{mora2019RAL} to account for the robot's nontrivial dimensions and $R^{\rm d}(\varphi_{\nu,k})$ is the 2D rotation matrix
\begin{equation}\nonumber
    R^{\rm d}(\varphi_{\nu,k}) = \begin{bmatrix}
     \cos(\varphi_{\nu,k}) & -\sin(\varphi_{\nu,k}) \\
    \sin(\varphi_{\nu,k}) & \phantom{-}\cos(\varphi_{\nu,k})
    \end{bmatrix} \enspace.
\end{equation}

\section{Details on Numerical Experiments}
\label{app:numerical-experiments}

The \ac{MPC}s are solved using MATLAB's $\mathsf{fmincon}$ with all simulations performed on an intel i7 PC with 16 GB RAM. Note that we have not considered dynamic obstacles in the numerical simulations detailed in Section~\ref{sec:MPC_Perf}. 

\subsubsection{\ac{LIP} parameters}
The \ac{LIP} parameters are selected to (approximately) match the geometry of Digit as follows. The step duration is $T = 0.3 \mathrm{s}$ and the height is $H = 0.91 \rm{m}$. 

\subsubsection{Reachability constraints}
The reachability constraints in \eqref{eq:constraint_set} are specified by  $lb^{x_c}_k = -0.2\rm{m}$ and $ub^{x_c}_k = 0.5 \rm{m}$ for both the left and right support, $lb^{y_c}_k = -0.5 \rm{m}$ and $ub^{y_c}_k = -0.2 \rm{m}$ for the left and $lb^{y_c}_k = 0.2 \rm{m}$ and $ub^{y_c}_k = 0.5 \rm{m}$ for the right support. We select $lb^{\theta} = -15^\circ$ and $ub^{\theta} = 15^\circ$ and $\delta_{\rm min} = 0\mathrm{m}$ and $\delta_{\rm max} = 0.2\mathrm{m}$. 

\subsubsection{Weighting matrices}
The weighting matrices in \eqref{eq:MPC_terminal} and \eqref{eq:MPC_running} are 
\begin{align} 
\nonumber
W_1 &= \mathsf{diag} \{5, 5, 2, 10, 10\} \\
\nonumber
W_2 &= \mathsf{diag} \{0.5, 0.5, 2, 10, 10 \}\\
\nonumber
W_3 &= \mathsf{diag}\{30, 30, 30, 30, 30\} \enspace.
\end{align}

\subsubsection{\ac{CBF} parameters}
The \ac{CBF} parameters in \eqref{eq:CBF_constraint} and \eqref{eq:barrier_moving} are $\gamma = 0.1$ and $\gamma_{\nu} = 0.2$ respectively. 

\section{Details on the MPC Implementation on Digit}
\label{app:Digit-MPC}

\begin{figure}[b]
    \centering
    \includegraphics{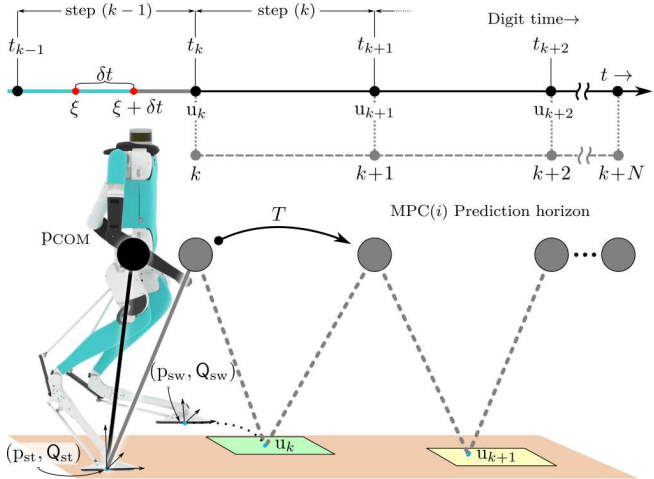}
    \vspace{-0.2in}
    \caption{\ac{MPC} implementation on Digit. The \ac{MPC} is solved at time $\xi$ during step $(k-1)$th step. Its solution $\mathrm{u}_k$ becomes available at $\xi+\delta t$ and is used to update the swing foot configuration at touchdown. Red dots on the continuous time $t$ axis are the times when the \ac{LIP} state, $\mathrm{x}_\mathsf{init}$ is extracted from the Digit to predict $\mathrm{x}_k$ as required by the $\mathrm{MPC}$. This prediction is based on the solution \eqref{eq:LIP_x_continuous}-\eqref{eq:LIP_theta_continuous} of the continuous dynamics of the \ac{LIP}.    }
    \label{fig:timing}
\end{figure}

With reference to Fig.~\ref{fig:timing}, let $[t_{k-1}, t_k)$ be the time interval over which step $(k-1)$ of Digit takes place, and suppose that $\xi \in [t_{k-1}, t_k)$ is the instant at which we begin solving the \ac{MPC}; let $\delta t > 0$ be the time required to compute a solution (see Fig.~\ref{fig:timing}). The initial condition $\mathrm{x}_\mathrm{init}$ required by the \ac{MPC} is predicted as follows. At $\xi$, feedback from Digit's state is used to provide initial conditions for the continuous-time dynamics of the 3D-\ac{LIP}. For example, in \eqref{eq:LIP_x_continuous} we set\footnote{Notation: the subscripts $\square_\mathrm{COM}$  $\square_\mathrm{sw}$ and $\square_\mathrm{st}$ denote quantities corresponding to Digit's \ac{COM}, and swing and stance foot, respectively.} $x(0) = \mathrm{p}^x_\mathrm{COM}(\xi)$, $\dot{x}(0) = \dot{\mathrm{p}}^x_\mathrm{COM}(\xi)$ and $u^x = \mathrm{p}^x_\mathrm{COM}(\xi) - \mathrm{p}^x_\mathrm{st}(\xi)$, and we compute $x(t)$ and $\dot{x}(t)$ at $t = kT-\xi$, which is the time left until the \ac{LIP}-predicted end of the current step. Similarly for the dynamics along the $y$ axis. Finally, $\theta(0)$ is set equal to the yaw angle $\phi^z_\mathrm{st}$ of the foot in contact with the ground during step $(k-1)$; since $\dot{\theta}=0$, we have $\theta(t) = \phi^z_\mathrm{st}$ at $t = kT-\xi$. With this information, we define $\mathrm{x}_\mathrm{init} = \begin{bmatrix} x(t) & \dot{x}(t) & y(t) & \dot{y}(t) & \theta(t) \end{bmatrix}^\mathsf{T}_{t=kT-\xi}$ and the corresponding \ac{MPC} is solved and the solution is available after time $\delta t$.

\section{Details on the OSC-QP Controller}
\label{app:OSC-QP}

A quaternion, $\mathsf{Q}=\{\mathsf{Q}_\mathrm{s},\mathsf{Q}_\mathrm{v}\}$, is composed of a scalar $\mathsf{Q}_\mathrm{s} \in \mathbb{R}$ and a vector $\mathsf{Q}_\mathrm{v}\in \mathbb{R}^3$. The error  $\mathsf{E} : \mathbb{R}^4\times\mathbb{R}^4 \rightarrow \mathbb{R}^3$ between the desired $\mathsf{Q}^{\rm des} = \{\mathsf{Q}^{\rm des}_\mathrm{s}, \mathsf{Q}^{\rm des}_\mathrm{v}\}$ and actual $\mathsf{Q} = \{\mathsf{Q}_\mathrm{s}, \mathsf{Q}_\mathrm{v}\}$ orientation in quaternion coordinates is defined as
\begin{equation}\nonumber
\mathsf{E}(\mathsf{Q}^{\rm des},\mathsf{Q})=\mathsf{Q}_\mathrm{s}^{\rm des}\mathsf{Q}_\mathrm{v} - \mathsf{Q}_\mathrm{s}\mathsf{Q}_\mathrm{v}^{\rm des}+\mathsf{Q}_\mathrm{v}^{\rm des}\times \mathsf{Q}_\mathrm{v}\enspace.
\end{equation}
With this definition we can now define the error $\eta_1$ between the desired and actual trajectories used to compute the cost function $\Psi_1$ as follows
\begin{equation} \nonumber
    \eta_1(s) = 
    \begin{bmatrix}
        \mathsf{E}(\mathsf{Q}_\mathrm{b}^\mathrm{des}(s), \mathsf{Q}_\mathrm{b}(s))\\
        \mathrm{p}_\mathrm{COM}^\mathrm{des}(s) - \mathrm{p}_\mathrm{COM}(s)\\
        \mathsf{E}(\mathsf{Q}_\mathrm{sw}^\mathrm{des}(s), \mathsf{Q}_\mathrm{sw}(s))\\
        \mathrm{p}_\mathrm{sw}^\mathrm{des}(s)-\mathrm{p}_\mathrm{sw}(s)
    \end{bmatrix}_{12\times 1} \enspace.
\end{equation}

The parameters associated with the \ac{OSC}-\ac{QP} controller are given below. The gains used to compute $\Psi_1$ are 
\begin{align}
\nonumber
K_\mathrm{P} =\mathsf{diag}\{&25, 25, 6.25, 25, 25,\\
\nonumber
&100, 2500, 2500, 25, 625, 625, 225\}    
\end{align}
and 
\begin{equation}\nonumber
K_{\mathrm{D}} = 2\sqrt{K_\mathrm{P}} \enspace.
\end{equation}
The gain matrix associated with $\Psi_2$ is
\begin{equation}\nonumber
K_\mathrm{G} = \mathsf{diag}\{1, 1, 1\} \enspace.
\end{equation}
In $\Psi_3$, the desired configuration for the left and right arms is
\begin{equation}\nonumber
    \eta_{\rm des} = {\begin{bmatrix} -0.1 & 0.9 & 0 & 0.4 & 0.1 & -0.9 & 0 & -0.4 \end{bmatrix}}^{\mathsf{T}} \mathrm{rad}
\end{equation}
and the corresponding gains are $K_{\mathrm{P,a}} = 100 I_{8 \times 8}$ and $K_{\mathrm{D,a}} = 10 I_{8 \times 8}$, where $I$ denotes the identity matrix.
Finally, the weights in the OSC-QP are $\psi_1 = 100$, $\psi_2 = 3$, $\psi_3 = 0.3$ and $\psi_4 = 0.1$.

\end{appendices}

\end{document}